\documentclass[10pt,twocolumn,letterpaper]{article}

\usepackage{cvpr}

\usepackage[dvipsnames]{xcolor}

\definecolor{cvprblue}{rgb}{0.21,0.49,0.74}
\usepackage{bm}
\usepackage{verbatim}
\usepackage{amsmath}
\usepackage{amssymb}
\usepackage{booktabs}
\usepackage{float}
\usepackage{makecell}
\usepackage{xcolor}
\usepackage[pagebackref,breaklinks,colorlinks,citecolor=cvprblue]{hyperref}
\usepackage[accsupp]{axessibility}

\def\paperID{15867} 
\def\confName{CVPR}
\def\confYear{2024}

\title{DeCoTR: Enhancing Depth Completion with 2D and 3D Attentions}

\author{
Yunxiao Shi~~~
Manish Kumar Singh~~~
Hong Cai~~~
Fatih Porikli~~~
\smallskip
\\
Qualcomm AI Research$^*$
\\
\smallskip
{\tt\small\{yunxshi, masi, hongcai, fporikli\}@qti.qualcomm.com}
}

\newcommand{\ours}{{DeCoTR}\xspace}

\begin{document}
\maketitle
\begin{abstract}

In this paper, we introduce a novel approach that harnesses both 2D and 3D attentions to enable highly accurate depth completion without requiring iterative spatial propagations. Specifically, we first enhance a baseline convolutional depth completion model by applying attention to 2D features in the bottleneck and skip connections. This effectively improves the performance of this simple network and sets it on par with the latest, complex transformer-based models. Leveraging the initial depths and features from this network, we uplift the 2D features to form a 3D point cloud and construct a 3D point transformer to process it, allowing the model to explicitly learn and exploit 3D geometric features. In addition, we propose normalization techniques to process the point cloud, which improves learning and leads to better accuracy than directly using point transformers off the shelf. Furthermore, we incorporate global attention on downsampled point cloud features, which enables long-range context while still being computationally feasible. We evaluate our method, DeCoTR, on established depth completion benchmarks, including NYU Depth V2 and KITTI, showcasing that it sets new state-of-the-art performance. We further conduct zero-shot evaluations on ScanNet and DDAD benchmarks and demonstrate that DeCoTR has superior generalizability compared to existing approaches.

\end{abstract}    
\section{Introduction}
\label{sec:intro}
\vspace{-3pt}

{\let\thefootnote\relax\footnotetext{{
\hspace{-6.5mm} $^*$ Qualcomm AI Research is an initiative of Qualcomm Technologies, Inc.}}}

Depth is crucial for 3D perception in various downstream applications, such as 
 autonomous driving, augmented and virtual reality, and robotics~\cite{chen2015deepdriving, bai2020depthnet, diaz2017designing, du2020depthlab, deheuvel2023learning,shi2023ega,zhu1232020mda,zhu2019structure,shi2019pairwise,yasarla2023mamo,shi2019self}. However, sensor-based depth measurement is far from perfect. Such measurements often exhibit sparsity, low resolution, noise interference, and incompleteness. Various factors, including environmental conditions, motion, sensor power constraints, and the presence of specular, transparent, wet, or non-reflective surfaces, contribute to these limitations. 
Consequently, the task of depth completion, aimed at generating dense and accurate depth maps from sparse measurements alongside aligned camera images, has emerged as a pivotal research area~\cite{cheng2018depth, ma2018sparse, cheng2019learning, xu2019depth, lin2022dynamic, park2020nonlocal, liu2022graphcspn, youmin2023completionformer}.

Thanks to the advances in deep learning, there has been significant progress in depth completion. Earlier papers leverage convolutional neural networks to perform depth completion with image guidance and achieve promising results~\cite{ma2018sparse, chen2019learning, tang2020learning}. In order to improve accuracy, researchers have studied various spatial propagation methods~\cite{liu2017learning,cheng2018depth,park2020nonlocal, lin2022dynamic}, which performs further iterative processing on top of depth maps and features computed by an initial network. Most existing solutions build on this in the last stage of their depth completion pipeline to improve performance~\cite{hu2021penet, youmin2023completionformer}. These propagation algorithms, however, focus on 2D feature processing and do not fully exploit the 3D nature of the problem. A few recent papers utilize transformers for depth completion~\cite{youmin2023completionformer, rho2022guideformer}. However, they apply transformer operations mainly to improve feature learning on the 2D image plane and fail to achieve acceptable accuracy without employing spatial propagation.

Several studies have looked into harnessing 3D representation more comprehensively. For instance, \cite{huynh2021boosting, zhou2023bev} construct a point cloud from the input sparse depth, yet coping with extreme sparsity poses challenges in effective feature learning. Another approach, as seen in \cite{liu2022graphcspn}, uplifts 2D features to 3D by using the initial dense depth predicted by a simple convolutional network, but it is impeded by the poor accuracy of the initial network and requires dynamic propagations to attain acceptable accuracy. Very recently, researchers have proposed employing transformers for 3D feature learning in depth completion~\cite{yu2023aggregating}; however, this work applies transformer layers to extremely sparse points, which is ineffective for learning informative 3D features.
\begin{figure*}
    \centering
    \includegraphics[width=0.98\textwidth]{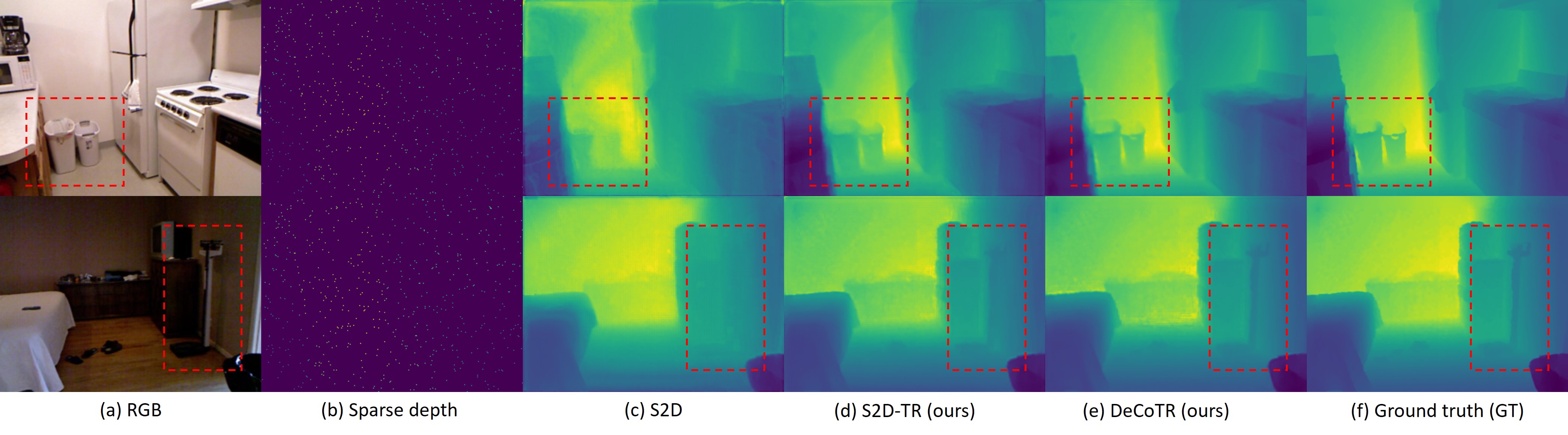}
    \vspace{-8pt}
    \caption{\small Example depth completion results on NYU Depth v2 dataset~\cite{silberman2012indoor}. We first upgrade S2D (c) to S2D-TR with efficient attention on 2D features, which significantly improves the initial depth completion accuracy (d). Based on the more accurate initial depths, \ours uplifts 2D features to form a 3D point cloud and leverages cross-attention on 3D points, which leads to highly accurate depth completion, with sharp details and close-to-GT quality (e). We highlight sample regions where we can clearly see progressively improving depths by using our proposed designs.}
    \label{fig:teaser}
    \vspace{-10pt}
\end{figure*}

Here, we introduce DeCoTR to perform feature learning in full 3D. It accomplishes this by constructing a dense feature point cloud derived from completed depth values obtained from an initial network and subsequently applying transformer processing to these 3D points. To do this properly, it is essential to have reasonably accurate initial depths. As such, we first enhance a commonly used convolution-based initial depth network, S2D~\cite{ma2018sparse}, by integrating transformer layers on bottleneck and skip connection features. This upgraded model, termed \emph{S2D-TR}, achieves significantly improved depth accuracy, on par with state-of-the-art models, without requiring any iterative spatial propagation.

Given the initial depth map, we uplift 2D features to 3D to form a point cloud, which is subsequently processed by transformer layers, to which we refer as \emph{3D-TR} layers. Prior to feeding the points to transformer layers, we normalize them, which regularizes the 3D feature learning and leads to better accuracy. In each 3D-TR layer, we follow standard practice~\cite{zhao2021point, wu2022point} to perform neighborhood-based attention, as global attention would be computationally intractable when the number of points is large. To facilitate long-range contextual understanding, we additionally incorporate global attention on lower-scale versions of the point cloud. Finally, 3D features are projected back to the 2D image plane and consumed by a decoder to produce the final depth prediction. As we shall see in the paper, our proposed transformer-based learning in full 3D provides considerably improved accuracy and generalizability for depth completion; see Fig.~\ref{fig:teaser} for a visual example.



In summary, our main contributions are as follows:
\begin{itemize}
    \item We present \ours, a novel transformer-based approach to perform full 3D feature learning for depth completion. This enables high-quality depth estimation without requiring iterative processing steps. 
    \item In order to properly do this, we upgrade the commonly used initial network S2D, by enhancing its bottleneck and skip connection features using transformers. The resulting model, S2D-TR, performs on-par with SOTA and provides more correct depths to subsequent 3D learning.
    \item We devise useful techniques to normalize the uplifted 3D feature point cloud, which improves the model learning. We additionally apply low-resolution global attention to 3D points, which enhances long-range understanding without making computation infeasible.
    \item Through extensive evaluations on standard benchmarks, NYU Depth v2~\cite{silberman2012indoor} and KITTI~\cite{geiger2013vision}, we demonstrate the efficacy of \ours and show that it sets the new SOTA, e.g., new best result on NYU Depth v2. Our zero-shot testing on ScanNet~\cite{dai2017scannet} and DDAD~\cite{guizilini20203d} further showcases the better generalizability of our model as compared to existing methods.
\end{itemize}


\begin{figure*}
    \centering
    \includegraphics[width=0.98\textwidth]{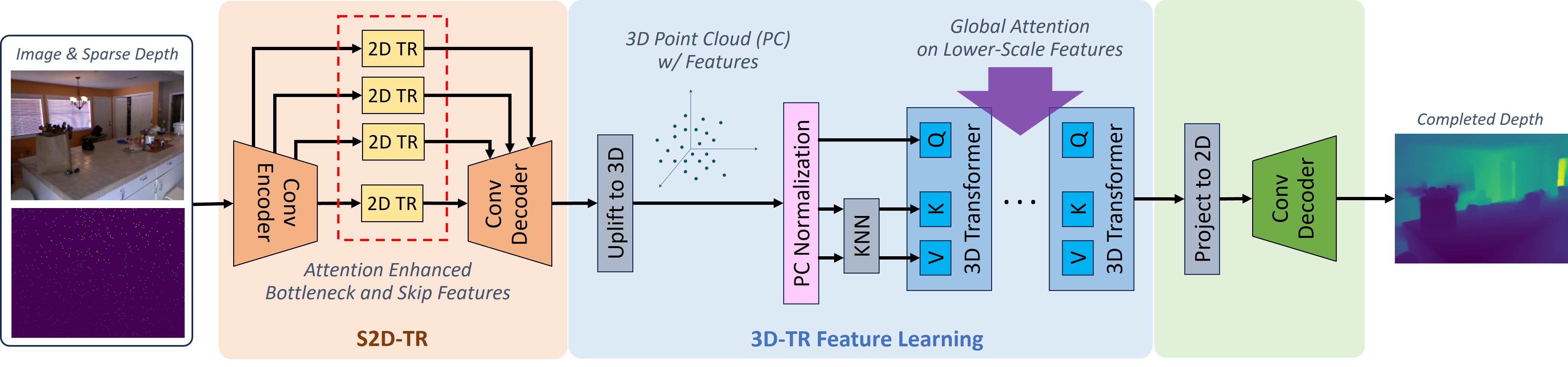}
    \vspace{-8pt}
    \caption{Overview of our proposed DeCoTR. The input RGB image and sparse depth map are first processed by our S2D-TR, which upgrades S2D with efficient 2D attentions. The learned 2D guidance features from S2D-TR are then uplifted to form a 3D feature point cloud based on the initial completed depth map. We normalize the point cloud and feed it through multiple 3D cross-attention layers (3D-TR) to enable geometry-aware feature learning and processing. We also introduce efficient global attention to capture long-range scene context. The attended 3D features from 3D-TR are projected back to 2D and given to a decoder to output the final completed depth map.}
    \label{fig:diagram}
    \vspace{-10pt}
\end{figure*}

\section{Related Works}
\label{sec:related}
\vspace{-3pt}

\textbf{Depth completion:} 
Early depth completion approaches~\cite{uhrig2017sparsity, hawe2011dense, ku2018defense} rely solely on the sparse depth measurements to estimate the dense depth. Since these methods do not utilize the image, they usually suffer from artifacts like blurriness, especially at object boundaries. 
Later, image-guided depth completion alleviates these issues by incorporating the image. S2D~\cite{ma2018sparse}, one of the first papers on this, leverages a convolutional network to consume both the image and sparse depth map. Subsequent papers design more sophisticated convolutional models for depth completion~\cite{qiu2019deeplidar,chen2019learning, imran2021depth, zhao2021adaptive, tang2020learning}. 
In order to enhance depth quality, researchers have studied various spatial propagation algorithms~\cite{cheng2018depth, cheng2019learning, park2020nonlocal, lin2022dynamic}. These solutions utilize depth values and features given by an initial network (usually S2D), and performs iterative steps to mix and aggregate features on the 2D image plane. In many papers nowadays, it has become a common practice to use spatial propagation on top of the proposed depth completion network in order to achieve state-of-the-art accuracy~\cite{hu2021penet, nazir2022semattnet, youmin2023completionformer}. Some recent works more tightly integrate iterative processing into the network, using architectures like recurrent network~\cite{wang2023lrru} and repetitive hourglass network~\cite{yan2022rignet}.

While existing solutions predominately propose architectures to process features on 2D, several works explore 3D representations. 
For instance, \cite{huynh2021boosting, zhou2023bev, yu2023aggregating} considers the sparse depth as a point cloud and learn features from it. However, the extremely sparse points present a challenge to feature learning. One of these works, GraphCSPN~\cite{liu2022graphcspn}, employs S2D as an initial network to generate the full depth map, before creating a denser point cloud and performing feature learning on it. However, this is limited by the insufficient accuracy of the initial depths by S2D and still needs iterative processing to achieve good accuracy.

\vspace{2pt}
\textbf{Vision transformer:} 
Even since its introduction~\cite{dosovitskiy2020image}, vision transformers have been extensively studied and utilized for various computer vision tasks, including classification, detection, segmentation, depth estimation, tracking, 3D reconstruction, and more. We refer readers to these surveys for a more comprehensive coverage of these works. More related to our paper are those that leverage vision transformers for depth completion, such as CompletionFormer~\cite{youmin2023completionformer} and GuideFormer~\cite{rho2022guideformer}. While they demonstrate the effectiveness of using vision transformers for depth completion, their feature learning is only performed on the 2D image plane. A very recent paper, PointDC~\cite{yu2023aggregating}, proposes to apply transformer to 3D point cloud in the depth completion pipeline. However, PointDC operates on very sparse points, which makes it challenging for learning 3D features.
\section{Method}
\label{sec:method}
\vspace{-3pt}
In this section, we present our proposed approach, \ours, powered by efficient 2D and powerful 3D attention learning. The overall pipeline of \ours is shown in Fig.~\ref{fig:diagram}. 

\subsection{Problem Setup}\vspace{-3pt}
Given aligned sparse depth map ${S}\in\mathbb{R}^{H\times W}$ and an RGB image ${I}\in\mathbb{R}^{H\times W\times 3}$, the goal of image-guided depth completion is to recover a dense depth map ${D}\in\mathbb{R}^{H\times W}$ based on $S$ and with semantic guidance from $I$. The underlying reasoning is that visually similar adjacent regions are likely to have similar depth values. 
Formally, we have \vspace{-5pt}
\begin{equation}
    {D} = H({S}, {I}),
\vspace{-5pt}
\end{equation}
where $H$ is a depth completion model to be learned.

It is a common approach to employ early fusion between the depth and RGB modalities~\cite{ma2018sparse,park2020nonlocal,liu2022graphcspn}. This has the advantage of enabling features to contain both RGB and depth information early on, so that the model can learn to rectify incorrect depth values by leveraging neighboring, similar pixels that have correct depths. We follow the same practice, first encoding RGB ${I}$ and sparse depth ${S}$ with two separate convolutions to obtain image and depth features $f_{I}$ and $f_{S}$ respectively: \vspace{-5pt}
\begin{equation}
    f_{I} = \text{conv}_{rgb}(I),\quad f_{S} = \text{conv}_{dep}(S)
\vspace{-5pt}
\end{equation}
which are then concatenated channel-wise to generate the initial fused feature, $f_1\in\mathbb{R}^{C_1\times H_1\times W_1}$.

\subsection{Enhancing Baseline with Efficient 2D Attention}\label{2dtr}
\vspace{-3pt}
The early-fusion architecture of S2D~\cite{ma2018sparse} has been commonly used by researchers as a base network to predict an initial completed depth map (e.g.,~\cite{park2020nonlocal, liu2022graphcspn}). Given the initial fused feature $f_1$, S2D continues to encode $f_1$ and generates multi-scale features $f_m\in\mathbb{R}^{C_m\times H_m\times W_m}$ for $m = \{2,...,5\}$, where $C_m$, $H_m$, $W_m$ are the number of channels, and height and width of the feature map. A conventional decoder with convolutional and upsampling layers is used to consume these features, where the smallest feature map is directly fed to the decoder and larger ones are fed via skip connections. The decoder has two prediction branches, one producing a completed depth map and the other generates guidance feature $g$. 

This architecture, however, has limited accuracy and may provide erroneous depth values for subsequent operations, such as 2D spatial propagation or 3D representation learning. As such, we propose to leverage self-attentions to enhance the S2D features. More specifically, we apply Multi-Headed Self-Attention (MHSA) to each $f_m$. Since MHSA incurs quadratic complexity in both time and memory w.r.t to input resolution, in order to avoid the intractable costs of processing high-resolution feature maps, we first employ depth-separable convolutions~\cite{chollet2017xception} to summarize large feature maps and reduce their resolutions to the same size (in terms of height, width, and channel) as the smallest feature map. The downsized feature maps are denoted as $\tilde{f}_m\in\mathbb{R}^{C_k\times H_k\times W_k}$, $\forall m>1$. 
Three linear layers are used to derive query $\tilde{q}_i\in\mathbb{R}^{N_k\times C_k}$, key $\tilde{k}_m \in\mathbb{R}^{N_k\times C_k}$, value $\tilde{v}_m\in\mathbb{R}^{N_k\times C_k}$ for each $\tilde{f}_m$, where $N_k=H_k\cdot W_k$. Next, we apply self-attention for each $\tilde{f}_m$:\vspace{-5pt}
\begin{equation}
    \tilde{f}^{A}_m = \text{softmax}\Big(\frac{\tilde{q}_m\tilde{k}_m^{\top}}{\sqrt{C_k}}\Big) \tilde{v}_m,
\vspace{-5pt}
\end{equation}
where $\tilde{f}^{A}_m$ denotes the attended features. These features are restored to their original resolutions by using depth-separable de-convolutions and we denote the restored versions as $f^A_m$. Finally, we apply a residual addition to obtain the enhanced feature map for each scale:\vspace{-5pt}
\begin{equation}
    f^E_m = f^{A}_m + f_m,\quad \forall m=\{2,...,5\}.
\vspace{-5pt}
\end{equation}

The enhanced feature maps are then consumed by the decoder to predict the initial completed depth map and guidance feature map. Our upgraded version of S2D with efficient attention enhancement, denoted as S2D-TR, provides significantly improved accuracy while having better efficiency than latest transformer-based depth completion models. For instance, S2D-TR achieves a lower RMSE (0.094 vs. 0.099) with $\sim$50\% less computation as compared to CompletionFormer without spatial propagation~\cite{youmin2023completionformer}.

\subsection{Feature Cross-Attention in 3D}
\label{3dtr}
\vspace{-3pt}
Considering the 3D nature of depth completion, it is important for the model to properly exploit 3D geometric information when processing the features. 
To enable this, we first un-project 2D guidance feature from S2D-TR, based on the initial completed depth, to form a 3D point cloud. This is done as follows, assuming a pinhole camera model with known intrinsic parameters:\vspace{-5pt}
\begin{equation}
    \begin{bmatrix}
        x\\
        y\\
        z\\
        1
    \end{bmatrix}=d
    \begin{bmatrix}
        1/\gamma_u & 0 & -c_u/\gamma_u & 0\\
        0 & 1/\gamma_v & -c_v/\gamma_v & 0\\
        0 & 0 & 1 & 0\\
        0 & 0 & 0 & 1
    \end{bmatrix}
    \begin{bmatrix}
        u\\
        v\\
        1\\
        1/d
    \end{bmatrix}
\vspace{-5pt}
\end{equation}
where $\gamma_u, \gamma_v$ are focal lengths, $(c_u, c_v)$ is the principal point, $(u, v)$ and $(x, y, z)$ are the 2D pixel coordinates and 3D coordinates, respectively. 

Given the large number of 3D points uplifted from the 2D feature map, it is computationally intractable to perform attention on all the points simultaneously. As such, we adopt a neighborhood-based attention, by finding the K-Nearest-Neighboring (KNN) points for each point in the point cloud, ${p}_i\in\mathbb{R}^3$, which we denote as $\mathcal{N}(i)$.


To bake 3D geometric relationship into the feature learning process, we perform cross-attention between the feature of each point and features of its neighboring points. 
Concretely, we modify from the original point transformer~\cite{zhao2021point,wu2022point} to implement this. 
For each point $p_i$, linear projections are first applied to transform its guidance feature ${g}_i$ to query ${q}_i$, key ${k}_i$, and value ${v}_i$. Following~\cite{zhao2021point}, we use vector attention which creates attention weights to modulate individual feature channels. More specifically, the 3D cross-attention is performed as follows:\vspace{-5pt}
\begin{align}\label{eq:vattn}
    & {a}_{ij} = w(\phi({q}_i, {k}_j)),\\
    & {g}_i^{a} = \sum_{j\in\mathcal{N}(i)}\text{softmax}({A}_i)_j\odot{v}_j,
\vspace{-5pt}
\end{align}
where $\phi$ is a relation function to capture the similarity between a pair of input point features (we use subtraction here), $w$ is a learnable encoding function that computes attention scores to re-weight the channels of the value,  ${A}$ is the attention weight matrix whose entries are ${a}_{ij}$ for points $p_i$ and $p_j$, ${g}_i^A$ denotes the output feature after cross-attention for $p_i$, and $\odot$ denotes the Hadamard product. We perform such 3D cross-attention in multiple transformer layers, to which we refer as 3D-TR layers.

While it is possible to directly use existing point transformers off-the-shelf, we find that this is not optimal for depth completion. Specifically, we incorporate the following technical designs to improve the 3D feature learning process. 

\textbf{Point cloud normalization:} We normalize the constructed point cloud from S2D-TR outputs into a unit ball, before proceeding to the 3D attention layers. We find this technique effectively improves depth completion, as we shall show in the experiments.

\textbf{Positional embedding:} Instead of the positional embedding multiplier proposed in~\cite{wu2022point}, we adopt the conventional one based on relative position difference. We find the more complex positional embedding multiplier does not benefit the learning and incurs additional computational cost.

\subsection{Capturing Global Context in 3D}
\vspace{-3pt}
The 3D cross-attention discussed previously updates each point feature only based on the point's estimated 3D neighborhood, in order to maintain computation tractability given the quadratic complexity of attention w.r.t. number of points. 
However, global or long-range scene context is also important for the model to develop accurate 3D understanding. 
To enable global understanding while keeping computation costs under control, we propose to perform global 3D cross-attention only on a downsampled point set, at the last encoding stage of the point transformer. In this case, we use the scalar attention as follows: \vspace{-5pt}
\begin{equation}
     {g}_i^{ga} = \sum_{j\neq i}\text{softmax}\Big(\frac{\langle{q}_i, {k}_j\rangle}{\sqrt{C_g}}\Big){v}_j,
\vspace{-5pt}
\end{equation}
where $\langle\cdot\rangle$ denotes dot product and $C_g$ is the embedding dimension. We apply global attention after the local neighborhood-based attentions.



\subsection{Training}\vspace{-3pt}
We train \ours with a masked $\ell_1$ loss between the final completed depth maps and the ground-truth depth maps, following standard practice as in \cite{park2020nonlocal,liu2022graphcspn}. More formally, the loss is given by\vspace{-5pt}
\begin{equation}
    \mathcal{L}(D^{gt}, D^{pred}) = \frac{1}{N}\sum_{i,j}\mathbb{I}_{\{d_{i,j}^{gt}>0\}}\Big|d_{i,j}^{gt} - d_{i,j}^{pred}\Big|,
\vspace{-5pt}
\end{equation}
where $\mathbb{I}$ is the indicator function, $d_{i,j}^{gt}\in D^{gt}$ and $d_{i,j}^{pred}\in D^{pred}$ represent pixel-wise depths in ground-truth and predicted depth maps, respectively,  $N$ is the total number of valid pixels.

\begin{table*}
\small
  \centering
  \begin{tabular}{lccccc}
    \toprule
    \textbf{Method} & RMSE $\downarrow$ & Abs Rel $\downarrow$ & $\delta < 1.25$ $\uparrow$ & $\delta < 1.25^2$ $\uparrow$ & $\delta < 1.25^3$ $\uparrow$ \\
    \midrule
    S2D~\cite{ma2018sparse} & 0.204 & 0.043 & 97.8 & 99.6 & 99.9 \\
    DeepLiDAR~\cite{qiu2019deeplidar} & 0.115 & 0.022 & 99.3 & 99.9 & 100.0 \\
    CSPN~\cite{cheng2018depth} & 0.117 & 0.016 & 99.2 & 99.9 & 100.0 \\
    DepthNormal~\cite{xu2019depth} & 0.112 & 0.018 & 99.5 & 99.9 & 100.0 \\
    ACMNet~\cite{9440471} & 0.105 & 0.015 & 99.4 & 99.9 & 100.0 \\
    GuideNet~\cite{tang2020learning} & 0.101 & 0.015 & 99.5 & 99.9 & 100.0 \\
    TWISE~\cite{imran2021depth} & 0.097 & \underline{0.013} & 99.6 & 99.9 & 100.0 \\
    NLSPN~\cite{park2020nonlocal} & 0.092 & \textbf{0.012} & 99.6 & 99.9 & 100.0 \\
    RigNet~\cite{yan2022rignet} & 0.090 & \underline{0.013} & 99.6 & 99.9 & 100.0 \\
    DySPN~\cite{lin2022dynamic} & 0.090 & \textbf{0.012} & 99.6 & 99.9 & 100.0 \\
    CompletionFormer~\cite{youmin2023completionformer} & 0.090 & \textbf{0.012} & - & - & - \\
    \midrule
    PRNet~\cite{lee2021depth} & 0.104 & 0.014 & 99.4 & 99.9 & 100.0 \\
    CostDCNet~\cite{10.1007/978-3-031-20086-1_15} & 0.096 & \underline{0.013} & 99.5 & 99.9 & 100.0 \\
    PointFusion~\cite{huynh2021boosting} & 0.090 & 0.014 & 99.6 & 99.9 & 100.0 \\
    GraphCSPN~\cite{liu2022graphcspn}  & 0.090 & \textbf{0.012} & 99.6 & 99.9 & 100.0 \\
    PointDC~\cite{yu2023aggregating} & 0.089 & \textbf{0.012} & 99.6 & 99.9 & 100.0 \\
    \midrule
    \ours (ours) & \underline{0.087} & \textbf{0.012} & 99.6 & 99.9 & 100.0\\
    \ours w/ GA (ours) & \textbf{0.086} & \textbf{0.012} & {99.6} & {99.9 } &{100.0}\\
    \bottomrule
  \end{tabular}
  \vspace{-6pt}
  \caption{Quantitative evaluation of depth completion performance on NYU-Depth-v2. GA denotes global attention.  RMSE and REL are in meters. Methods in the top part of the table focus on feature learning and processing in 2D and those in the bottom block exploit 3D representation. Best and second best numbers are highlighted in bold and underlined, respectively, for RMSE and Abs Rel.}
  \label{tab:nyud}
  \vspace{-0pt}
\end{table*}

\section{Experiments}
\label{sec:experiments} \vspace{-3pt}
We conduct extensive experiments to evaluate our proposed \ours on standard depth completion benchmarks and compare with the latest state-of-the-art (SOTA) solutions. We further perform zero-shot evaluation to assess the model generalizability and carry out ablation studies to analyze different parts of our proposed approach.

\subsection{Experimental Setup}\vspace{-3pt}

\noindent \textbf{Datasets:} We perform standard depth completion evaluations on NYU Depth v2 (NYUD-v2)~\cite{silberman2012indoor} and KITTI Depth Completion (KITTI-DC)~\cite{geiger2012we, geiger2013vision}, and generalization tests on ScanNet-v2~\cite{dai2017scannet} and DDAD~\cite{guizilini20203d}. These datasets cover a variety of indoor and outdoor scenes. We follow the sampling settings from existing works to create input sparse depth~\cite{park2020nonlocal, liu2022graphcspn}.

NYUD-v2 provides RGB images and depth maps captured by a Kinect device from 464 different indoor scenes. We use the official split: 249 scenes for training and the remaining 215 for testing. Following the common practice~\cite{park2020nonlocal,liu2022graphcspn,youmin2023completionformer}, we sample $\sim$50,000 images from the training set and resize the image size from $480\times 640$ first to half and then to $228\times 304$ with center cropping. We use the official test set of 654 images for evaluation.

KITTI is a large real-world dataset in the autonomous driving domain, with over 90,000 paired RGB images and LiDAR depth measurements. There are two versions of KITTI dataset used for depth completion. One is from~\cite{ma2018sparse}, which consists of 46,000 images from the training sequences for training and a random subset of 3,200 images
from the test sequences for evaluation. The other one is KITTI Depth Completion (KITTI-DC) dataset, which provides 86,000 training, 6,900 validation, and 1,000 testing samples with corresponding raw LiDAR scans and reference images. We use KITTI-DC to train and test our model on the official splits.

ScanNet-v2 contains 1,513 room scans reconstructed from RGB-D frames. The dataset is divided into 1,201 scenes for training and 312 for validation, and provides an additional 100 scenes for testing. For sparse input depths, we sample point clouds from vertices of the reconstructed meshes. We use the 100 test scenes to evaluate depth completion performance, with 20 frames randomly selected per scene. We remove samples where more than 10\% of the ground-truth depth values are missing, resulting in 745 test frames across all 100 test scenes.

DDAD is an autonomous driving dataset collected in the
U.S. and Japan using a synchronized 6-camera array, featuring long-range (up to 250m) and diverse urban driving scenarios. Following~\cite{guizilini20203d}, we downsample the images from the original resolution of $1216\times 1936$ to $384\times 640$. We use the official 3,950 validation samples for evaluation. Since after downsampling there's only less than $5\%$ valid ground truth depth, for our method and all the comparing we sample all the available valid depth points so that reasonable results are generated.



\noindent \textbf{Implementation Details:} 
We implement our proposed approach using PyTorch~\cite{paszke2019pytorch}. We use the Adam~\cite{kingma2014adam} optimizer with an initial learning rate of $5\times 10^{-4}$, $\beta_1=0.9$, $\beta_2=0.999$, and no weight decay. The batch size for NYUDv2 and KITTI-DC per GPU is set to 8 and 4,  respectively. All experiments are conducted on 8 NVIDIA A100 GPUs. 

\vspace{3pt}
\noindent \textbf{Evaluation:} 
We use standard metrics to evaluate depth completion performance~\cite{eigen2014depth}, including Root Mean Squared Error (RMSE), Absolute Relative Error (Abs Rel), $\delta\!<\!1.25$, $\delta\!<\!1.25^2$, and $\delta\!<\!1.25^3$. On KITTI-DC test, we use the official metrics: RMSE, MAE, iRMSE and iMAE. 
We refer readers to the supplementary file for detailed mathematical definitions of these metrics. The depth values are evaluated
with maximum distances of 80 meters and 200 meters for KITTI
and DDAD, respectively, and 10 meters for NYUD-v2 and ScanNet.

\subsection{Results on NYUD-v2 and KITTI}\vspace{-3pt}



\begin{figure*}[t!]
    \centering
    \includegraphics[width=0.98\textwidth]{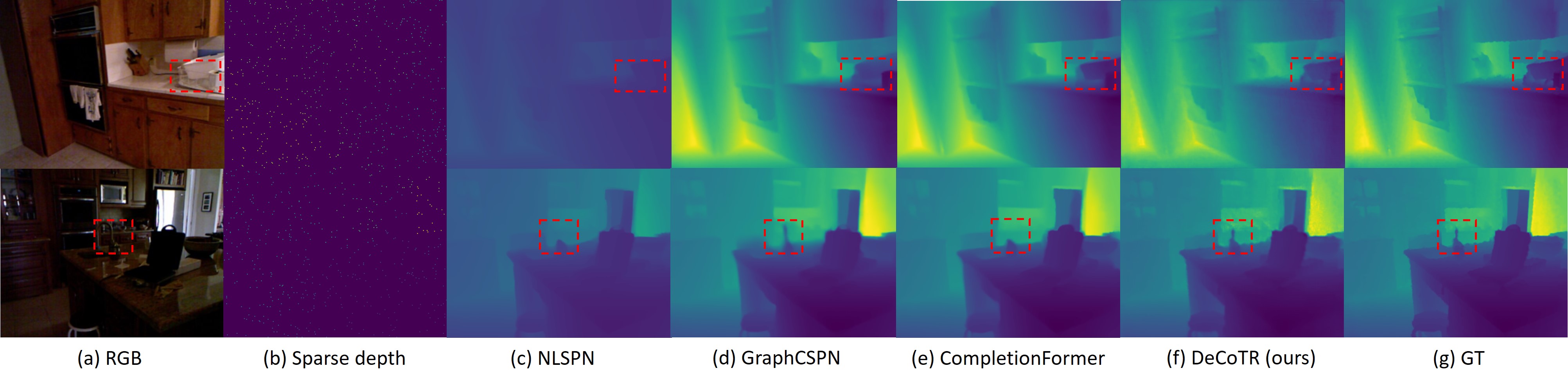}
    \vspace{-10pt}
    \caption{Qualitative results on NYUD-v2. We compare with SOTA methods such as NLSPN, GraphCSPN, and CompletionFormer. Areas where \ours provides better depth accuracy are highlighted.}
    \vspace{-10pt}
    \label{fig:nyu}
\end{figure*}

\textbf{On NYUD-v2:} Table~\ref{tab:nyud} summarizes the quantitative evaluation results on NYUD-v2. Our proposed \ours approach sets the new SOTA performance, with the lowest RMSE of 0.086 outperforming all existing solutions. When not using 3D global attention, \ours already provides the best accuracy and global attention further improves it. 
Specifically, our \ours considerably outperforms latest SOTA methods that also leverage 3D representation and/or transformers, such as GraphCSPN, PointDC, and CompletionFormer. Note that although PointDC uses both 3D representation and transformer, it only obtains slightly lower RMSE when comparing to methods that do not use 3D or transformer (e.g., CompletionFormer, GraphCSPN). This indicates that the PointDC approach is suboptimal, potentially due to the extremely sparse 3D points.

Fig.~\ref{fig:nyu} provides sample qualitative results on NYUD-v2. We see that \ours generates highly accurate dense depth maps that are very close to the ground truth. The depth maps produced by \ours capture much finer details as compared to existing SOTA methods. For instance, in the second example, our proposed approach accurately predicts the depth on the faucet despite its small size in the images and the low contrast, while other methods struggle.

\textbf{On KITTI-DC:} We evaluate \ours and compare with existing methods (including latest SOTA) on the official KITTI test set, as shown in Table~\ref{tab:kitti}. \ours achieves SOTA depth completion accuracy and is among the top-ranking methods on KITTI-DC leaderboard.\footnote{Top-5 among published methods at the time of submission, in terms of iRMSE, iMAE, and MAE.} We see that \ours performs significantly better than existing SOTA methods that leverage 3D representations, e.g., GraphCSPN, PointDC. This indicates that \ours has the right combination of dense 3D representation and transformer-based learning. 

\begin{figure*}
    \centering
    \includegraphics[width=0.95\textwidth]{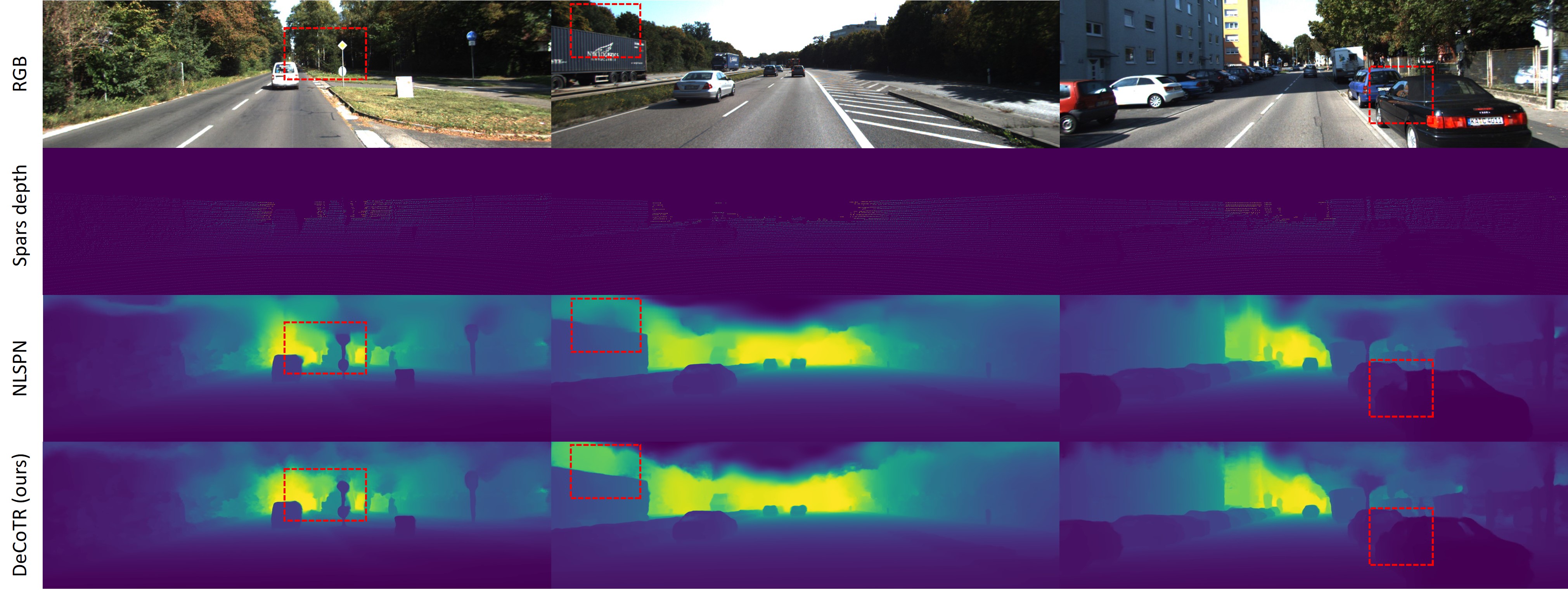}
    \vspace{-15pt}
    \caption{Qualitative results on KITTI DC. Areas where \ours provides better depth accuracy are highlighted.}
    \label{fig:kitti}
    \vspace{-11pt}
\end{figure*}

Fig.~\ref{fig:kitti} shows visual examples of our completed depth maps on KITTI. \ours is able to generate correct depth prediction where NLSPN produces erroneous depth values; see the highlighted areas in the figure. For instance, in the second example, \ours accurate estimates the depth around the upper edge of the truck while the depth map by NLSPN is blurry in that region.

\begin{table*}
\small
  \centering
  \begin{tabular}{lccccc}
    \toprule
    \textbf{Method} & RMSE $\downarrow$ & MAE $\downarrow$ & iRMSE $\downarrow$ & iMAE $\downarrow$ \\
    \midrule
    CSPN~\cite{cheng2018depth} & 1019.64 & 279.46 & 2.93 & 1.15 \\
    TWISE~\cite{imran2021depth} & 840.20 & 195.58 & 2.08 & \textbf{0.82} \\
    ACMNet~\cite{9440471} & 744.91 & 206.09 & 2.08 & 0.90 \\
    GuideNet~\cite{tang2020learning} & 736.24 & 218.83 & 2.25 & 0.99 \\
    NLSPN~\cite{park2020nonlocal} & 741.68 & 199.59 & 1.99 & \underline{0.84} \\
    PENet~\cite{hu2021penet}  & 730.08 & 210.55 & 2.17 & 0.94 \\
    GuideFormer~\cite{rho2022guideformer} & 721.48 & 207.76 & 2.14 & 0.97 \\
    RigNet~\cite{yan2022rignet} & 712.66 & 203.25 & 2.08 & 0.90 \\
    DySPN~\cite{lin2022dynamic} & \underline{709.12} & \textbf{192.71} & \textbf{1.88} & \textbf{0.82} \\
    CompletionFormer~\cite{youmin2023completionformer} & \textbf{708.87} & 203.45 & 2.01 & 0.88 \\
    \midrule
    PRNet~\cite{lee2021depth} & 867.12 & 204.68 & 2.17 & 0.85 \\
    FuseNet~\cite{chen2019learning} & 752.88 & 221.19 & 2.34 & 1.14 \\
    PointFusion~\cite{huynh2021boosting} & 741.9 & 201.10 & 1.97 & 0.85 \\
     GraphCSPN~\cite{liu2022graphcspn}  & 738.41 & 199.31 & 1.96 & \underline{0.84} \\
    PointDC ~\cite{yu2023aggregating} & 736.07 & 201.87 & 1.97 & 0.87 \\
    \midrule
    \ours (ours) & 717.07 & \underline{195.30} & \underline{1.92} & \underline{0.84} \\
    \bottomrule
  \end{tabular}
  \vspace{-6pt}
  \caption{Quantitative evaluation of depth completion performance on official KITTI Depth Completion test set. RMSE and MAE are in millimeters, and iRMSE and iMAE are in 1/km. Similar to Table~\ref{tab:nyud}, methods in the top part focus on feature learning in 2D and those in the bottom block exploit 3D representation. Best and second best numbers are highlighted in bold and underlined, respectively.}
  \label{tab:kitti}
  \vspace{-4pt}
\end{table*}

\subsection{Zero-Shot Testing on ScanNet and DDAD}\vspace{-3pt}
Most existing papers only evaluate their models on NYUD-v2 and KITTI, without looking into model generalizability. In this part, we perform cross-dataset evaluation. More specifically, we run zero-shot testing of NYUD-v2-trained models on ScanNet-v2 and KITTI-trained models on DDAD. This will allow us to understand how well our \ours as well as existing SOTA models generalize to data not seen in training.

Tables~\ref{tab:scannet} and~\ref{tab:ddad} present evaluation results on ScanNet-v2 and DDAD, respectively. We see that \ours generalizes better to unseen datasets when comparing to existing SOTA models. It it noteworthy to mention that on DDAD, \ours has significantly lower depth errors as compared to both NLSPN and CompletionFormer, despite that CompletionFormer has slightly lower RMSE on KITTI-DC test. Moreover, in this case, CompletionFormer has even worse accuracy than NLSPN, indicating its poor generalizability.
\vspace{-5pt}

\begin{figure*}
    \centering
    \includegraphics[width=0.98\textwidth]{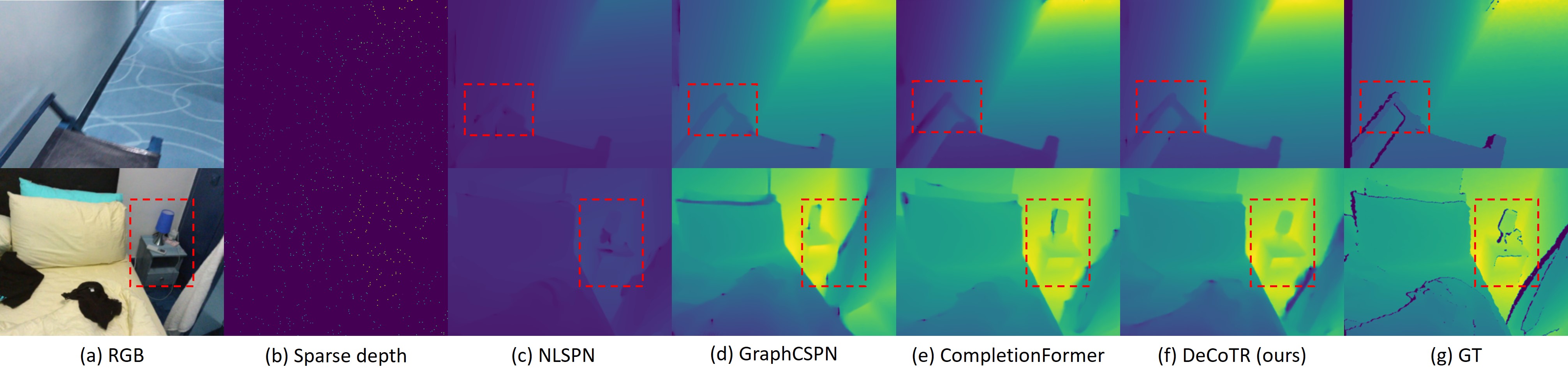}
    \vspace{-10pt}
    \caption{Qualitative results of zero-shot inference on ScanNet-v2. Areas where \ours provide better depth accuracy are highlighted.}
    \vspace{-6pt}
    \label{fig:scannet}
\end{figure*}

\begin{figure}[t]
    \centering
    \includegraphics[width=0.4\textwidth]{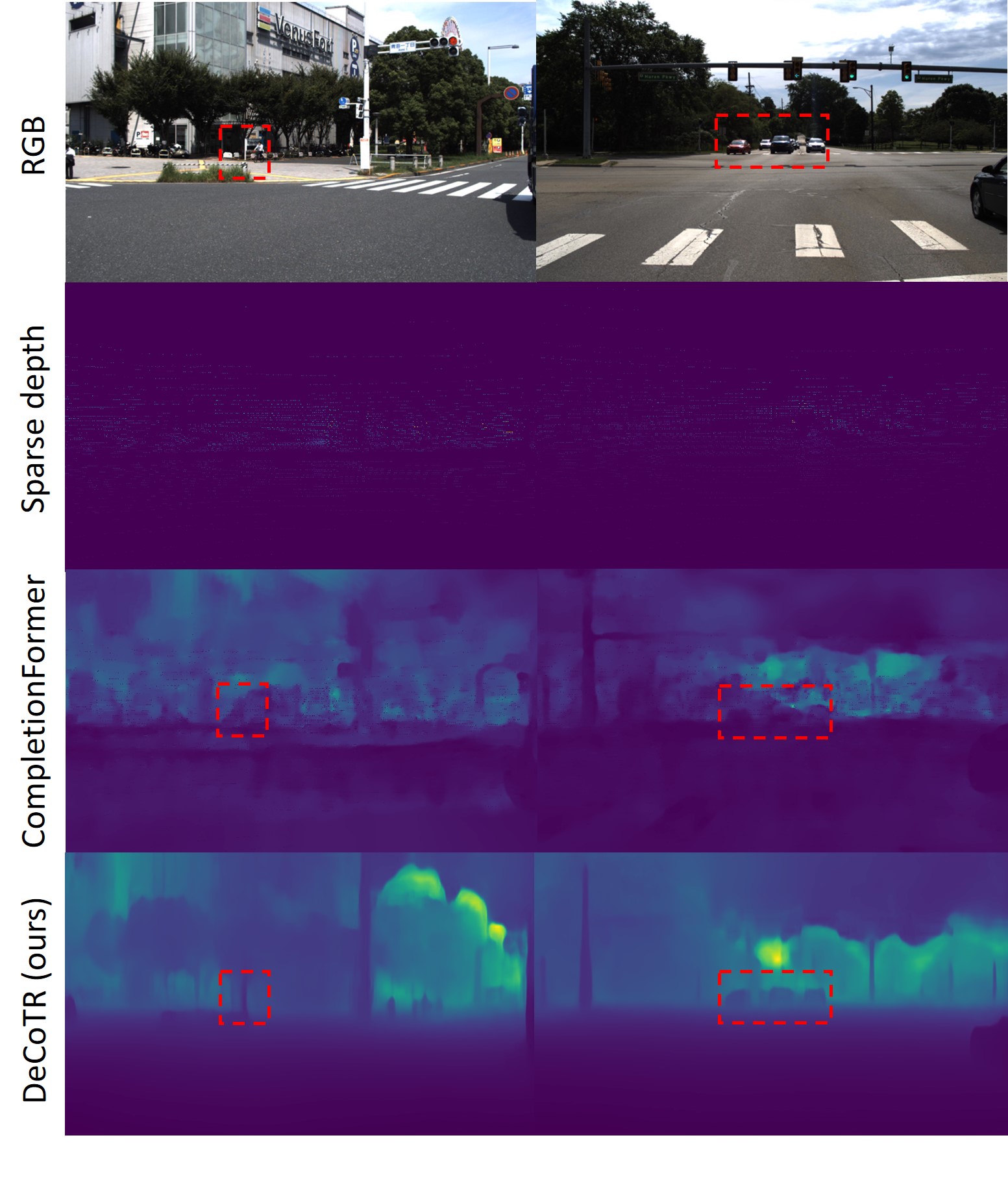}
    \vspace{-12pt}
    \caption{Qualitative results of zero-shot inference on DDAD. Areas where \ours provide better depth accuracy are highlighted.}
    \label{fig:ddad}
    \vspace{-10pt}
\end{figure}

Fig.~\ref{fig:scannet} shows sample visual results of zero-shot depth completion on ScanNet-v2. \ours generates highly accurate depth maps and captures fine details, e.g., arm rest in the first example, lamp in the second example. Other methods cannot recover the depths accurately. Fig.~\ref{fig:ddad} provides qualitative results on DDAD for CompletionFormer and our \ours. While this is a challenging test setting given the much larger depth range in DDAD, \ours still predicts reasonable depths. In contrast, it can be seen that CompletionFormer performs very poorly on DDAD. We notice that \ours's predictions are more accurate in the nearer range (e.g., on cars) and less so when it is far away (e.g., on trees), since KITTI training only covers up to 80 meters whereas DDAD has depth up to 200 meters. This is also confirmed by the lower-than-KITTI RMSE and higher-than-KITTI MAE numbers of \ours on DDAD.

\begin{table}[t!]
\small
  \centering
  \begin{tabular}{lccccc}
    \toprule
    \textbf{Method}  & RMSE $\downarrow$ & $\delta < 1.25$ $\uparrow$  \\
    \midrule
    NLSPN~\cite{park2020nonlocal}& 0.198 & 97.3  \\
    GraphCSPN~\cite{liu2022graphcspn} & 0.197 & 97.3  \\
    CompletionFormer~\cite{youmin2023completionformer}  & 0.194 & 97.3  \\
    \midrule
    {\ours} (ours) & \textbf{0.188} & \textbf{97.6}  \\
    \bottomrule
  \end{tabular}
  \vspace{-8pt}
  \caption{Zero-shot testing on ScanNet-v2 using models trained on NYUD-v2. Best numbers are highlighted in bold.}
  \label{tab:scannet}
  \vspace{-2pt}
\end{table}

\begin{table}
\small
  \centering
  \begin{tabular}{lccccc}
    \toprule
    \textbf{Method} & RMSE $\downarrow$ & MAE $\downarrow$ \\
    \midrule
    NLSPN~\cite{park2020nonlocal} & 701.9  &309.6 \\
    CompletionFormer~\cite{youmin2023completionformer} &889.3  &400.1  \\
    \midrule
    \ours (ours) &399.2   &263.1 \\
    \bottomrule
  \end{tabular}
  \vspace{-8pt}
  \caption{Zero-shot testing on DDAD using models trained on KITTI. Best numbers are highlighted in bold.}
  \label{tab:ddad}
  \vspace{-8pt}
\end{table}


\subsection{Ablation Study}\vspace{-3pt}
In this part, we investigate the effectiveness of various design aspects of our proposed \ours solution. Table~\ref{tab:abl} summarizes the ablation study results. Starting from the S2D baseline, we significantly improve depth completion performance by introducing efficient attention on the 2D features, reducing RMSE from 0.204 to 0.094. Next, by using neighborhood-based cross-attention on the 3D points (without normalizing the point cloud before 3D-TR layers), we reduce RMSE to 0.089. Even though scaling a 3D scene to a uniform perceived range may present a challenge to maintain the original spatial relationship, after applying our normalization scheme, \ours achieves a better RMSE of 0.087 and by additionally incorporating efficient global attention, the RMSE is further improved to 0.086. This study verifies the usefulness of our proposed components and techniques.

Note that if we directly apply 3D-TR on top of the original S2D network (second row in the table), we can still drastically improve upon S2D but fail to outperform existing methods that leverage 3D or transformers such as GraphCSPN and CompletionFormer. This confirms the importance of getting more accurate initial depth before applying 3D feature learning.

\section{Conclusion}
\label{sec:conclusion} \vspace{-3pt}
In this paper, we proposed a novel approach, \ours, for image-guided depth completion, by employing transformer-based learning in full 3D. We first proposed an efficient attention scheme to upgrade the common baseline of S2D, allowing S2D-TR to provide more accurate initial depth completion. 2D features are then uplifted to form a 3D point cloud followed by 3D-TR layers that apply powerful neighborhood-based cross-attention on the 3D points. We further devised an efficient global attention operation to provide scene-level understanding while keeping computation costs in check. 
Through extensive experiments, we have shown that \ours achieves SOTA performance on standard benchmarks like NYUD-v2 and KITTI-DC. Furthermore, zero-shot evaluation on unseen datasets such as ScanNet and DDAD shows that \ours has better generalizability as compared to existing methods.

\begin{table}[t]
\small
  \centering
  \begin{tabular}{lccc}
    \toprule
    \textbf{Method} & RMSE $\downarrow$ & $\delta < 1.25$ $\uparrow$  \\
    \midrule
    S2D~\cite{ma2018sparse} & 0.204 & 97.8 \\
    \midrule
    S2D-TR  & 0.094  & 99.4 \\
    S2D + 3D-TR &0.091&  99.6 \\
    \ours w/o normalization & 0.089 & 99.6 \\
    \ours  &  0.087 & 99.6 \\
    \ours w/ global attention &  0.086 & 99.6  \\
    \bottomrule
  \end{tabular}
  \vspace{-8pt}
  \caption{Ablation study conducted on NYUD-v2.}
  \label{tab:abl}
  \vspace{-8pt}
\end{table}
{
    \small
    \bibliographystyle{ieeenat_fullname}
    \bibliography{main}
}


\end{document}